\pdfoutput=1

\documentclass[11pt]{article}

\usepackage{amsmath,amsfonts,bm}

\def\eqref#1{equation~\ref{#1}}

\def\1{\bm{1}}

\def\vc{{\bm{c}}}
\def\vd{{\bm{d}}}

\DeclareMathAlphabet{\mathsfit}{\encodingdefault}{\sfdefault}{m}{sl}
\SetMathAlphabet{\mathsfit}{bold}{\encodingdefault}{\sfdefault}{bx}{n}

\def\sA{{\mathbb{A}}}
\def\sB{{\mathbb{B}}}

\def\sD{{\mathbb{D}}}

\def\sS{{\mathbb{S}}}

\newcommand{\KL}{D_{\mathrm{KL}}}

\usepackage{emnlp2021}

\usepackage{times}
\usepackage{latexsym}

\usepackage[T1]{fontenc}

\usepackage[utf8]{inputenc}

\usepackage{microtype}

\usepackage{algorithm}
\usepackage{algorithmic}
\usepackage{microtype}
\usepackage{amsmath}
\usepackage{multirow}
\usepackage{amsmath,array,graphicx,amssymb}
\usepackage[export]{adjustbox}
\usepackage{subcaption}
\usepackage{multirow}
\usepackage{mathtools} 
\usepackage{tabularx}
\usepackage{breakcites}
\usepackage{xcolor}
\usepackage{array}
\usepackage{makecell}
\usepackage{tabularx}
\usepackage{mathrsfs}
\usepackage{scalerel}
\usepackage{soul}
\usepackage{colortbl}
\definecolor{mygray}{gray}{.9}
\usepackage{soul}

\usepackage{tikz}
\usetikzlibrary{trees}

\usepackage{makecell}
\usepackage{booktabs}

\usepackage{url}
\usepackage{subfiles}

\title{Refine and Imitate: Reducing Repetition and Inconsistency in Persuasion Dialogues via Reinforcement Learning and Human Demonstration}

\author{Weiyan Shi$^{1}$, Yu Li$^{2}$, Saurav Sahay$^{3}$, Zhou Yu$^{1}$\\
Columbia University$^1$ University of California, Davis$^2$, Intel Labs$^3$\\
\texttt{ws2634@columbia.edu, yooli@ucdavis.edu}\\
\texttt{saurav.sahay@intel.com, zy2461@columbia.edu}
}

\begin{document}
\maketitle
\begin{abstract}
Persuasion dialogue systems reflect the machine's ability to make strategic moves beyond verbal communication, and therefore differentiate themselves from task-oriented or open-domain dialogue systems and have their own unique values. However, the repetition and inconsistency problems still persist in dialogue response generation and could substantially impact user experience and impede the persuasion outcome. Besides, although reinforcement learning (RL) approaches have achieved big success in strategic tasks such as games, they require a sophisticated user simulator to provide real-time feedback to the dialogue system, which limits the application of RL on persuasion dialogues. To address these issues towards a better persuasion dialogue system, we apply RL to refine a language model baseline without user simulators, and distill sentence-level information about repetition, inconsistency, and task relevance through rewards. Moreover, to better accomplish the persuasion task, the model learns from human demonstration to imitate human persuasion behavior and selects the most persuasive responses. Experiments show that our model outperforms previous state-of-the-art dialogue models on both automatic metrics and human evaluation results on a donation persuasion task, and generates more diverse, consistent and persuasive conversations according to the user feedback. The code is available at \url{https://github.com/wyshi/consistency}.


\end{abstract}

\section{Introduction}
Persuasion dialogue systems have become an increasingly important subject in both social science and computational linguistics \cite{prakken2006formal,prakken2009models,wang2019persuasion,asai2020emotional}. Such systems aim to employ conversational strategies to change the audience's attitude or behaviour, 
and therefore, are inherently difficult to build with multiple challenges. The first one is that users often expect highly smooth conversation experience from persuasion systems in order to be persuaded \cite{shi2020effects}. So the long-standing problems of dialogue repetition and inconsistency can be especially salient in persuasion dialogue systems. Secondly, different from traditional dialogue tasks, the persuasion task is non-collaborative where the user and the system have different goals \citep{li2019missa}, and hence highly intellectual and strategic.


Previous studies have attempted to address the first challenge, the dialogue repetition and inconsistency problems, by changing the object function in supervised learning \cite{li2019inconsisent} or applying reinforcement learning (RL) \citep{li2016deep,liu2018dialogue}.  
But these methods either may lead to  uninterpretable behaviors, or rely on hand-crafted user simulators that are hard to design for persuasion dialogues. To tackle these challenges, we propose to extract a policy directly from the data and let the models learn from its own mistakes without the use of simulators. Leveraging decoding methods such as Nucleus Sampling \citep{NucleusSampling}, the finetuned language model can generate lexically diverse response candidates given the same context. 
Some candidates are appropriate, while others are repetitive or inconsistent. These good and bad examples are used as positive and negative feedback to the model through meaningful rewards in RL, and help refine the language model. 

Besides being diverse and consistent, a good response in persuasion dialogues also needs to accomplish the task: to persuade people. Existing work simply relied on the language models to generate persuasive responses \cite{li2019missa, wu2019alternating}, which could result in uncontrollable task-oblivious replies. To quantify intellectual persuasion activities, we employ imitation learning,  and ask human experts to demonstrate the persuasion process. We build a response imitator to imitate these human demonstrations and  select the most persuasive responses in our framework.

We evaluate our models on a donation persuasion task \citep{wang2019persuasion}, and deploy the persuasion systems on Amazon Mechanical Turk to interact with real users. The results on both automatic and human evaluations show that our systems achieve better persuasion outcomes (higher donation amount and donation ratio), and generates more diverse, consistent and persuasive responses compared to the baselines. 

This work  makes multiple contributions. Firstly, we propose the first RL-based persuasive dialogue system framework that achieves state-of-the-art performance on a complex donation persuasion task. Secondly, we design DialGAIL, an RL-based generative algorithm to refine a baseline language model for  dialogue  generation without the use of user simulators. Additionally, we introduce a human persuasion demonstration dataset that can be used for future research. 
Previous dialogue research has mostly focused on pure task-oriented dialogues and pure social conversations; but looking forward, it becomes more and more important to pay attention to strategic dialogues that involves both task and social components. We sincerely hope this work could inspire more research and discussions on strategic dialogues in the community. 



\section{Related Work}
Strategic dialogue tasks such as persuasion and negotiation have emerged and attracted more attention recently, given its wide applications in industry and daily life \cite{lewis2017deal,  he2018decoupling, wang2019persuasion, li2019missa, shi2020effects}. These tasks are close to human-human conversations and often contain both a specific task goal and social components to build rapport for better task completion. 
Previously,  \citet{mazzotta2007portia} proposed an agenda-based user-adapted persuasion system to build relationship with users and change their eating habit. \citet{yuan2008human} developed a dialogue system for educational debate with strategic heuristics. 
More recently, \citet{li2019missa} utilized large-scale language models to build a donation persuasion system by generating multiple responses and selecting appropriate candidates with human-defined rules. We take a similar approach to generate candidates but eliminate the manual work for rule design, and teach the model to select task-relevant candidates through human demonstration.

Although large-scale language models have achieved great success in many NLP tasks, these models still suffer from repetition and inconsistency when applied to dialogue tasks. 
Many previous studies have worked on these issues \cite{wu2019alternating, li2019inconsisent, song2020generate}. 
For example, \citet{li2019inconsisent} proposed to detect the inconsistency with natural language inference data, and penalize it with unlikelihood loss to achieve more consistent personality in open-domain dialogues. \citet{song2020generate} detected and rewrote the contradicting responses to achieve a more consistent personality. Our work tackles these problems with RL to reduce exposure bias in supervised learning and improve the interpretability. 

Previous work has also explored RL-based methods in dialogue system building \citep{li2016deep, liu2018dialogue, shi2019build, shi2019unsupervised}. For instance, \citet{li2016deep} integrated the goal of coherent into the reward design  towards more diverse dialogue generation. \citet{liu2018dialogue} presented a hybrid reinforcement and imitation learning approach to enable the agent to learn from interactions with users in task-oriented dialogues. 
However, such methods not only rely on hand-crafted user simulators that are inherently hard to build \citep{shi2019build} for persuasion systems, but also require meaningful rewards that  
are difficult to design. In this work, we propose to let the model learn from its own mistakes by generating multiple responses without the use of simulators. 

Our work is also closely related to response selection, which focuses on obtaining good context representations to match the context and 
retrieve the best response from a large collection of human-human conversations. However, such response selection models are highly dependent on the quality and availability of the underlying datasets. To address the \textit{data scarcity} issue, \citet{henderson2019training} pretrained a response selection model with large conversational corpora, and finetuned it on new domains in task-oriented settings for a better context representation. Instead of retrieving candidates from human dialogues, we adopt the imitation learning approach, and leverage language models' ability to generate coherent responses, and build a selector to imitate human selection process and choose among the generated candidates.

\begin{figure*}[htb!]
    \centering
    \includegraphics[width=\textwidth]{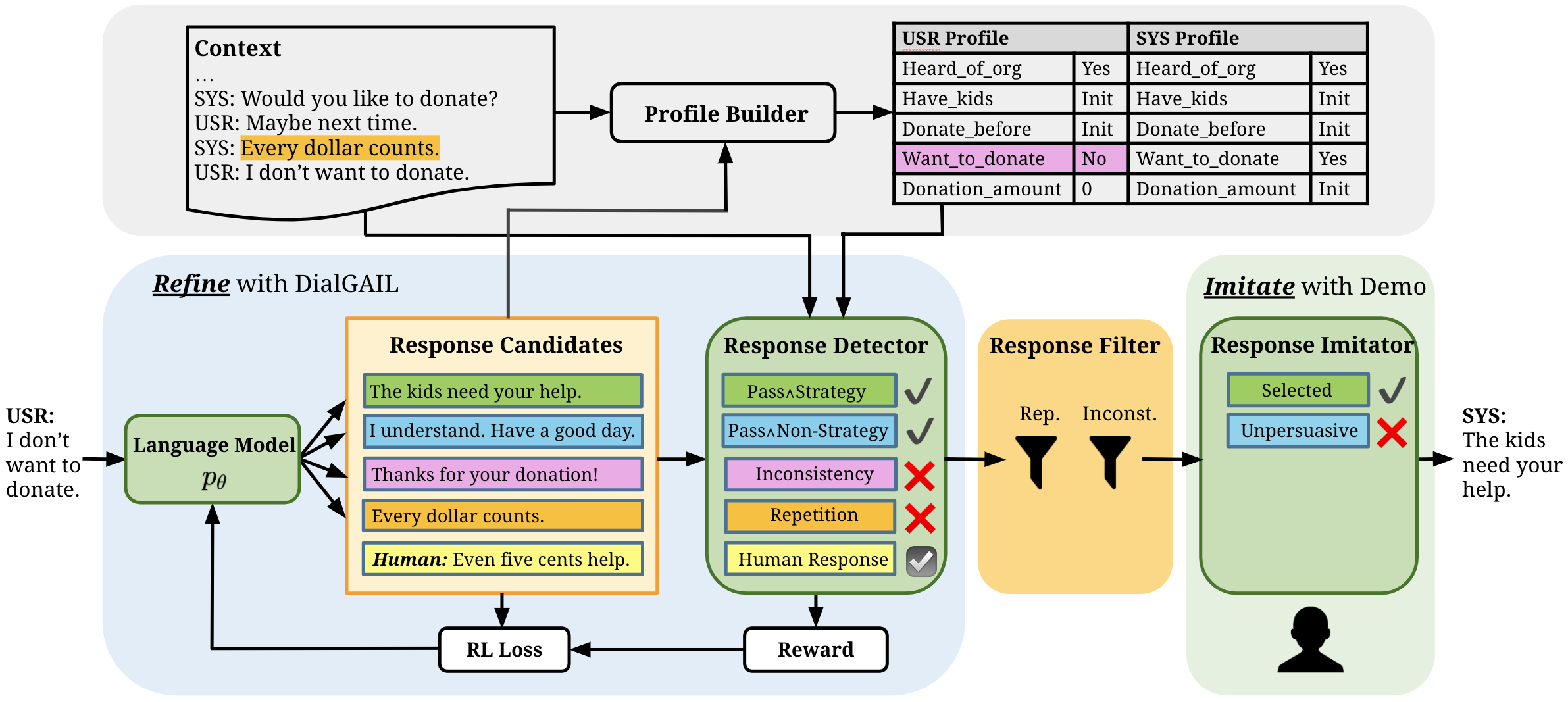}
    \caption{The overall architecture of our PersRFI model. During training, $p_{\theta}$ generates $n$ response candidates; \textit{Response Detector} annotates them with corresponding status such as ``Repetition''; and the response candidates along with the golden human response send feedback to refine $p_{\theta}$ through the rewards. During testing, the refined $p_{\theta^{\ast}}$ generates $n$ candidates again; \textit{Response Filter} removes the detected repetitive and inconsistent candidates; and \textit{Response Imitator} imitates human demonstrations to select the most persuasive candidate as the final output. The dialogue history consists of the dialogue context and the \textit{Profiles}.
    }
    \label{fig:model}
\end{figure*}

\section{Methods: PersRFI}
Our framework is shown in Figure~\ref{fig:model}. The language model is $p_{\theta}$ and there are two steps in the framework, 1) the reinforcement learning (RL) process to refine a baseline language  model  $q$ for better response generation (i.e., $p_{\theta_0}=q$), and 2) the imitation process to learn from human demonstration and select the best response. 
During RL training, for each user utterance, $p_{\theta}$ generates $n$ response candidates,  shown in the \textit{Response Candidates} box. Then the \textit{Response Detector} annotates these candidates with corresponding status such as ``Repetition'' and ``Inconsistency''. These labels along with the golden human response provide feedback through the reward function to guide $p_{\theta}$ to generate nonrepetitive and consistent responses. During test time, we use the refined language model $p_{\theta^\ast}$ to generate $n$ candidates again, and apply the \textit{Response Filter} to remove the repetitive and inconsistent candidates to further ensure the candidate quality. Finally, the \textit{Response Imitator} takes in the remaining candidates, and imitates the human demonstration to select one persuasive candidate as the final response. To detect repetition and inconsistency, we build \textit{USR Profile} and \textit{SYS Profile} shown in the top right table in Figure~\ref{fig:model}, where task-relevant information is extracted from the dialogue and stored as $<$\textit{key}: \textit{value}$>$ pairs, such as  ``\textit{want\_to\_donate: No}''.   
We describe each module below.

\subsection{Refine with Reinforcement Learning}
\subsubsection{DialGAIL}
One major issue with current RL-based dialogue training is that the it requires a sophisticated user simulator to provide real-time feedback to the dialogue system.  But in persuasion task, designing a persuadee simulator that can have diverse responses to persuasion is as hard as building the persuasion system itself. 
To eliminate the user simulator, we extend GAIL \cite{ho2016generative} to dialogues settings and propose DialGAIL. The basic idea is to start with a baseline model, then use it to explore more space by generating multiple responses, and finally provide different rewards to the responses to refine the original model. In this way, DialGAIL extracts a policy directly from the training dialogues and learn from its own mistakes.

Algorithm~\ref{algorithm} shows the steps in DialGAIL. We have a baseline model $q$ trained on the persuasion task, and initialize $p_{\theta}$ (the model being refined) with $q$. For each iteration, we sample one dialogue $\vd$ from the training corpus. For each turn in $\vd$, $p_{\theta}$ generates $n$ response candidates. Since persuasion strategies such as emotion appeal are found effective in human persuasion conversations \citep{wang2019persuasion}, to encourage more persuasion strategies, we classify the  candidates into ``Non-Strategy'' or ``Strategy'' with a dialogue-act classifier. Then the \textit{Response Detector} (described later) annotates each candidate with status $a_i\in$\{Human Response,  Pass$\land$Strategy, Pass$\land$Non-Strategy, Repetition, Inconsistency\}. 
With the detected status, candidates receive different rewards based on the  following conditions, 1) if it is a ground truth human response (highest reward), 2) if it contains persuasion strategy (medium reward), 3) if it is a repetitive or inconsistent response (lowest reward). The reward values are chosen based on the validation dataset performance and the reward function details for the donation task are in Section~\ref{sec:appendix, training details}. By optimizing the rewards, $p_{\theta}$ learns from its own repetitive and inconsistent mistakes and generates more diverse, consistent and persuasive responses. Note that although we choose repetition and inconsistency in our persuasion task, DialGAIL is not specific to reducing repetition and inconsistency only. Given corresponding response quality detectors, it can be generalized to improve other sentence-level qualities as well (e.g., naturalness, positive sentiment).  

\begin{algorithm}[ht!]
\caption{DialGAIL}

\begin{algorithmic}[1]
\STATE \textbf{Initialize}: Collect human-human dialogues $\sD$ \\ 
                        \qquad \qquad \,
                        Train $q$ with MLE on $\sD$\\
                            \qquad \qquad \,
                        Warm-up $p_{\theta}$ with $q$, i.e., $p_{\theta_0}=q$\\
                            \qquad \qquad \,
                            Initialize the Replay Buffer $\sB$
\FOR{i=$1,2,3, \dots$}
    \STATE Sample one dialogue $\vd$ from $\sD$
    \FOR{each turn in $\vd$} 
        \STATE $\vc$ = context, $s^{\ast}$ = human response
        \STATE  $p_{\theta_i}$ generates $n$ candidates $\sS=\{s_1, s_2,..., s_n\}$
        \STATE \textit{Response Detector} annotates $\sS$ with corresponding status $\sA=\{a_1, a_2,..., a_n\}$
        \STATE Put  the triplet  $(\vc, \{s^{\ast}\} \cup \sS, \{$``Human Response''$\}\cup \sA)$ into $\sB$;
        \STATE Continue the dialogue with $s^{\ast}$
    \ENDFOR
    \STATE Collect rewards for triplets in $\sB$
    \STATE Normalize the collected rewards
    \STATE Update $p_{\theta_i}$  with Eq.~(\ref{eq:generator}), and clear $\sB$
\ENDFOR
\end{algorithmic}

\label{algorithm}
\end{algorithm}

The next step is to train with DialGAIL. To stabilize the RL training process, we apply proximal policy optimization (PPO) \citep{schulman2017proximal} following \citet{wu2020textgail}. PPO performs importance sampling with the likelihood ratio between current and old policies $r(\theta) = \frac{p_{\theta_\text{i}}(s|\vc)}{p_{\theta_{\text{i-1}}}(s|\vc)}$,  and optimizes the surrogate in Eq.~(\ref{eq:surrogate}) to maximize the expected rewards.  To ensure the generation quality, we use the KL divergence between the  language model being refined $p_{\theta}$ and the baseline $q$ as the maximum entropy regularizer in RL. This KL-term prevents $p_{\theta}$ from moving too far away from the original model $q$ and potentially losing fluency. The final objective is shown  in Eq.~(\ref{eq:generator}), $s$ is the generated response and $s^{\ast}$ is the human response: 

\begin{equation}
\label{eq:surrogate}
\begin{aligned}
 L_\text{policy}(\theta) = \min &(r(\theta) \hat{A}_{s^{\ast}}, \\
                                &\text{clip}(r(\theta), 1-\epsilon, 1+\epsilon)\hat{A}_{s^{\ast}})) 
\end{aligned}
\end{equation}


\begin{equation}
    L(\theta) = \underset{s \sim p_\theta(\cdot|x)}{\mathbb{E}} [ L_\text{policy} (\theta) + \beta \, \KL ( q | \, p_\theta) ]
    \label{eq:generator}
\end{equation}





\subsubsection{Repetition and Inconsistency Detection}

\noindent \textbf{Profile Builder}. To apply DialGAIL, we need to detect the repetitive and inconsistent candidates. Previous methods treated this as a classification problem and required manual annotation of the inconsistency status \citep{welleck2018dialogue}. But manual annotations are expensive, and do not generalize across domains. Here we propose to build separate \textbf{Profiles} for \textit{both} the user and the system to track key contextual information  and detect the repetition and inconsistency more automatically. These profiles store $<$\textit{key}: \textit{value}$>$ pairs and are dynamically updated as the conversation unfolds. They are similar to \textit{dialogue state} in task-oriented dialogues, with the key difference that we track \textit{both} the system and the user in  strategic dialogue settings to avoid contradiction with the system's previous statements. In our task, experts analyze the human-human conversations and design an ontology with high-frequency questions such as ``Do you have kids'' (\textit{have\_kids}) as the keys in the profiles. For simplicity, we only track five attributes in the top grey table in Figure~\ref{fig:model}, but ideally new attributes should be added as the conversation continues and we leave this as future work. The \textit{Profile Builder} uses dialogue-act classifiers to build and update the  profiles. For example, if the last system-act is ``propose-donation'' and the following user-act is ``disagree-donation'', the user profile is updated with ``$<$\textit{want\_to\_donate}: \textit{No}$>$''.  The dialogue-act classifiers use GPT2-small and achieve 0.66 in F1 for system-act and 0.62 for user-act. 

\noindent \textbf{Repetition Detector.} One key observation is that  MLE-based baseline language models tend to repeat high-frequency sentences in the training corpus and usually repeat on the exact lexical level. Therefore, we calculate the Jaccard similarity coefficient between each context sentence $s_\text{ctx}$  and each candidate $s_\text{cdd}$, $\text{Ratio}_{\text{rep}} (s_\text{ctx}, s_\text{cdd}) = \frac{\text{Unigram}_{s_\text{ctx}}\cap \text{Unigram}_{s_\text{cdd}}}{\text{Unigram}_{s_\text{ctx}}\cup \text{Unigram}_{s_\text{cdd}}}$, as the repetition ratio after normalizing the text. If $\text{Ratio}_{\text{rep}}\geq0.5$, this candidate is considered as repetition. We experimented with other similarity metrics such as sentence embedding \citep{reimers2019sentence} and found that Jaccard similarity is the simplest but the most effective one without much computation overhead, because repetition usually happens on the lexical level in our persuasion task. Such simple detection is also task-independent and can be very easily generalized to other domains. In our final model, 9.0\% candidates are labeled as ``Repetition''. More details of the repetition detector are in the Appendix.


\noindent \textbf{Inconsistency Detector.} To detect inconsistency, we apply the \textit{Profile Builder} on each candidate, extract the value for each key, and  compare them against the current \textit{Profiles}. If the value extracted from the candidate contradicts the current \textit{Profiles}, it is detected as ``Inconsistency''. For example, the candidate ``Thanks for your donation'' in pink in Figure~\ref{fig:model} implies that the user \textit{want\_to\_donate:Yes}, which contradicts  \textit{want\_to\_donate:No} in the current \textit{USR Profile} and makes it an inconsistent candidate. In our experiments, 6.6\% candidates are inconsistent. We also trained a model on the Dialogue Natural Language Inference (DNLI) dataset \citep{welleck2018dialogue}  to detect inconsistency. However, the DNLI model's performance is limited, possibly because DNLI is annotated on the PersonaChat \citep{zhang2018personalizing}, which is very different from our persuasion task. We plan to explore domain-adaptation methods \citep{qian2019domain} to improve the inconsistency detector in the future.

\subsection{Response Filter}
Although DialGAIL has refined the language model, repetition and inconsistency can still happen due to the model's stochastic nature. Therefore, during testing time, we combine the repetition and inconsistency detectors to make a hard \textit{Response Filter} to filter out the bad candidates,  and send only the ``Pass'' candidates  to the next module. 
On average, 84.4\% candidates are ``Pass'' in our experiments. If no candidates pass the filter (i.e. out of candidates), the model will generate one additional sentence as the final response, which happened at a rate of only 0.2\% for our final model.

\subsection{Imitate with Human Demonstration}
Besides being nonrepetitive and consistent, a good response  also needs to move the conversation forward towards the task goal to persuade people to donate. However, intellectual activities such as persuasion or negotiation are difficult to quantify and optimize without imitation. Therefore, we perform behavior cloning \citep{bain1995framework} and ask humans to demonstrate the persuasion process for the model to imitate. One human expert was employed to interact with our model  for 10 conversations and was presented $n=$10 candidates for each turn. 
Since it is subjective to determine each candidate's persuasive level, to avoid bias towards different persuasive messages, the human expert was asked to select all acceptable responses given the context, rather than rating or ranking the candidates, which made the process easier and faster. In total, we collected 1,077 utterances (861 for training, 216 for validation) with binary labels (0 = not selected,  1 = selected) from the  expert, with the labor time being only 3 hours. We didn't employ more people in this process because we wanted to explore the potential of human demonstration. The experiments show that even with such small amount of data collection effort, human demonstration still helps significantly.

With the human demonstration data, we build the \textit{Response Imitator}, a binary classifier to imitate the human selection process. It takes in all ``Pass'' candidates that pass the \textit{Response Filter} and decide if a particular candidate is persuasive and should be selected.  
This classifier achieves 79.4\% in accuracy on the validation set. 
In our final model, 60.1\% candidates are selected.

It is worth noting that the \textit{Response Imitator} is fundamentally different from the ``next sentence prediction'' (NSP) classifier used in many  studies \citep{devlin2018bert, wolf2019huggingface}. Previous research shows that NSP doesn't help much in dialogue generation \citep{li2019missa}, partly because in NSP,  random sentences from the training data are assigned as negative examples. But in our response selection setting, the negative examples are generated by the language model under the same context, and therefore are semantically  much  closer to each other  and much harder to distinguish. This makes the \textit{Response Imitator} help more than the auxiliary NSP task in dialogue response generation, even with small amount of human effort.

\section{Experiments}

\begin{table*}[!htbp]

\centering
\resizebox{.85\textwidth}{!}{
\begin{tabular}{l|cccccc}
\midrule
 \textbf{Model} & \textbf{PPL} 
 & \textbf{OOC}   &  \textbf{Pass}   &  \textbf{Slct.} &  \textbf{Strag.} & \textbf{Len.}              \\ \midrule

MISSA \citep{li2019missa}   &19.91 &- &- &- &47.6\% & 16.62                     \\

ARDM \citep{wu2019alternating}& 12.45 & -     &  -      &     -     & 49.2\%         & 15.03
  \\
\midrule
PersRFI (Ours)  & \textbf{12.38} & \textbf{0.2\%}   & 84.4\%  & \textbf{60.1\%}    & \textbf{51.2\%}         & \textbf{19.36}*** 
  \\

PersRFI - RL (w/o RL)  & -     & 0.4\%   &   \textbf{85.3\%} & 59.2\%    & 49.6\%         & 18.29***    \\
PersRFI - RL - Demo (w/o RL w/o Demo)   & -     & 1.1\% &  83.9\% & -         & 41.5\%         & 15.12     
 \\

\midrule
                       
\end{tabular}
}
\caption{\textbf{Automatic evaluation results}. 
\textbf{OOC}: Out-of-candidate. \textbf{Pass}: Good candidates that are nonrepetitive and consistent and therefore pass the \textit{Response Filter}. \textbf{Slct.}: Persuasive candidates selected by the \textit{Response Imitator}. \textbf{Strag.}: Candidates with persuasion strategies.  The baselines only generate one response, so metrics that involve multiple  candidates such as OOC do not apply and are left blank. *p$<$0.05, **p$<$0.01. 
}
\label{tab:automatic}
\end{table*}

\subsection{Dataset}
We conduct our experiments on the \textsc{PersuasionForGood} dataset \citep{wang2019persuasion}. It has 1,017 rich human-human persuasion conversations, where one user persuades the other user to donate to \textit{Save the Children}\footnote{https://www.savethechildren.org/}. In the human-human setting, the average donation is \$0.35 with a persuadee donation probability of 0.54. Basic statistics of the dataset is shown in Table~\ref{tab:dataset statistic} in the Appendix.


\begin{table*}[!htbp]
\centering
\resizebox{0.9\textwidth}{!}{
\begin{tabular}{l|ccccc|cc}
\midrule
 \textbf{Model} & \textbf{Nonrep.} & \textbf{Const.} & \textbf{Fluc.} & \textbf{Pers.} &  \textbf{All.} &  \textbf{Dnt.}                   & \textbf{DntP.}           \\ \midrule

MISSA \citep{li2019missa}   & -       & 3.78       & 3.74     & -    & -    & \$0.41     & 0.50                    \\

ARDM \citep{wu2019alternating} & 3.17         & 3.95       & 4.17       & 2.33      & 3.61     & \$0.33 & 0.50 \\
\midrule
PersRFI (Ours)  &   3.50             &   \textbf{4.17}         &   \textbf{4.41}         &   \textbf{2.98}**       &4.0      &  \$0.53*      &    0.61\\

PersRFI - RL (w/o RL)  & \textbf{3.78}**       & 3.98       & 4.37     & 2.72     & \textbf{4.11}*    & \textbf{\$0.62}**     & \textbf{0.72}*                    \\
PersRFI - RL - Demo (w/o RL w/o Demo)   &   3.25        &    3.84    & 4.39       &   2.73  &    3.75   & \$0.38 & 0.57 \\

\midrule
                       
\end{tabular}
}
\caption{\textbf{Human evaluation results}.  \textbf{Nonrep.}: Nonrepetitiveness. \textbf{Const}: Consistency.  \textbf{Fluc.}: Fluency. \textbf{Pers.}: Persuasiveness. \textbf{All.}: Overall experience. \textbf{Dnt.}: Average donation. \textbf{DntP.}: Donation probability.   
*p$<$0.05, **p$<$0.01. 
}
\label{tab:human}

\end{table*}

\subsection{Baselines}



\noindent \textbf{MISSA} \citep{li2019missa}  is a transformer-based dialogue model \citep{wolf2019huggingface}  for strategic tasks  with human-designed response filters, 
 and jointly trains three tasks (language modeling, dialogue-act prediction and next sentence prediction).

\noindent \textbf{ARDM} \citep{wu2019alternating} uses two GPT2-medium models to model the user and the system separately, and jointly trains them to better capture different speakers' language styles. It achieves state-of-the-art results on the persuasion task, so we 
initialize $p_{\theta}$ with ARDM and refine it with DialGAIL.


\subsection{Evaluation Metrics}
We evaluate the models from two aspects:
\textbf{response quality} (measured by nonrepetitiveness, consistency, and fluency) and \textbf{persuasion outcome} (measured by persuasiveness, donation amount and donation probability). We conduct both automatic and human evaluations to assess the models.

\noindent \textbf{Automatic Metrics}. We use perplexity (PPL) to measure the models' generation quality. To evaluate the candidate quality, we  estimate the models' probability to run out of candidates (OOC),  the percentage of candidates that 1) are nonrepetitive and consistent and thus pass the \textit{Response Filter} (Pass); 2) are persuasive and selected by the \textit{Response Imitator} (Slct.);  3) have persuasion strategies (Strag.), and also the average sentence length (Len.).

\noindent \textbf{Human Evaluation}. We deployed the persuasive dialogue models on Amazon Mechanical Turk with ParlAI \citep{miller2017parlai} to interact with human users. Each model interacted with 50 unique users to persuade them to donate part of their task earnings to \textit{Save the Children}. Each user was allowed to do the task  only once to avoid bias. After the conversation, the users were asked to input their donation amount (Dnt.) privately, and rate the conversation on  nonrepetitiveness (Nonrep.), consistency (Const.), fluency (Fluc.), persuasiveness (Pers.),  and overall experience (All.) on five-scale.  Higher scores indicate better performances. We estimated the donation probability (DntP.) with the percentage of people who made a donation. 

\subsection{Quantitative Results}
The automatic and human evaluation results are shown in Table~\ref{tab:automatic} and \ref{tab:human} respectively. \textbf{PersRFI} refers to  our final model refined with DialGAIL (R) plus  \textit{Response Filter} (F) and \textit{Response Imitator} (I); \textbf{PersRFI - RL} refers to PersRFI minus refining with RL, which uses the baseline ARDM with the \textit{Response Filter} and the \textit{Response Imitator}. \textbf{PersRFI - RL - Demo} refers to PersRFI without RL refining and human demonstrations to train the  \textit{Response Imitator}, which is ARDM with the \textit{Response Filter} only. We performed one-tailed t-test between  ARDM and our three models.

In \textbf{automatic evaluation} in Table~\ref{tab:automatic}, we find that refining the model with DialGAIL achieves a lower perplexity (12.38 vs 12.45), indicating a better generation quality compared to the MISSA and ARDM baselines. PersRFI also generates more candidates with persuasion strategies than ARDM (51.2\% vs 49.2\%). Furthermore, PersRFI encourages longer generation and increases the average sentence length from 15.03 to 19.89 significantly. 

%
In \textbf{human evaluation} in Table~\ref{tab:human}, PersRFI outperforms all the baselines on all metrics. For response quality, it achieves the highest consistency score (4.17) and fluency score (4.41). For persuasion outcome, it also receives the highest persuasiveness score (2.98) with a  significantly higher average donation (\$0.53) than the baselines. The donation amount and donation probability are even higher than the human results in \textsc{PersuasionForGood} (average donation=\$0.35, donation probability=0.54).
We notice that the persuasiveness scores of all models are relatively low compared to other metrics, indicating that persuasion is indeed a hard task and worth studying.
All these results suggest that applying DialGAIL to refine the language model and imitating human demonstration to select the response are effective on all levels.

We report the \textbf{Ablation study} results in the lower half of Table~\ref{tab:automatic} and \ref{tab:human}, and find \textit{Response Filter} alone (PersRFI - RL - Demo) doesn't improve the model much, probably because the candidates that pass the filter are still randomly selected and therefore not persuasive. However, \textit{Response Imitator} makes significant contributions to reducing repetition and improving the overall experience, and also obtains the highest average donation amount (\$0.62) and the highest donation probability (0.72). 
This confirms that even small amount of human demonstrations can be very helpful in accomplishing complex tasks such as persuasion. Finally, adding RL further improves the model's persuasiveness (2.98 vs 2.72) and consistency (4.17 vs 3.98), decreases the out-of-candidate (OOC)  probability (0.2\% vs 0.4\%)  and leads to longer candidates (19.36 vs 18.29) with more strategies (51.2\% vs 49.6\%), indicating a better generation quality.

\begin{table*}[!htb]
\small
\centering
\begin{adjustbox}{width=0.95\textwidth}
\centering
\begin{tabular}{lm{150mm}}


\toprule

\multicolumn{2}{c}{

\normalsize \textbf{Model:} PersRFI 

}\\
\multicolumn{2}{l}{ 
\textbf{Rating:} nonrepetitiveness=5, consistency=5, fluency=5, persuasiveness=5, overall=5, donation=\$0.5.
}
\\


\hline
\rowcolor{mygray} SYS   &  Hi, how are you doing today?  \\\rowcolor{white}
USR   & hello, I'M WONDERFUL..And yourself?\\\rowcolor{mygray}
SYS   &  I am great, thank you. Have you ever heard of the charity Save the Children?  \\\rowcolor{white}
USR   & That GREAT... yes i have..\\\rowcolor{mygray}
SYS   &  It is an international non-governmental organization that promotes children's rights, provides relief and helps support children in developing countries.  \\\rowcolor{white}
USR   &That's great.. i love children organization.. They love to help many of family's.. \\\rowcolor{mygray}
SYS   &  Yes, they are also a very trustworthy organization. \ul{You are probably aware of the hurricane that hit Puerto Rico last week.} It was devastating for all of Puerto Rico and their children. Do you want to make a donation to Save the Children? You can choose any amount from \$0 to all your task payment (\$2) to donate, which will be directly deducted from your payment. \\\rowcolor{white}
USR   & I would love to help fund the organization.. but at the moment  i just started working... \\\rowcolor{mygray}
SYS   & \hl{They have a lot of resources available to them, from food and clothing to school supplies and medical attention.} If you want to know more about their programs and their unique approach to providing relief to children in need, they have a website  https://www.savethechildren.org/   \\\rowcolor{white}
USR   & okay.. Thank you i will follow-up .. Because i feel so bad that i cant give back right now..\\\rowcolor{mygray}
SYS   & You are very right. \hl{It is very easy to give to charity, but there is a lot of pressure on people to give more. Many people just feel that they are not good enough and are not worthy. They don't feel good about what they are doing, and feel like they don't have the right to make a difference.}  \\\rowcolor{white}
USR   & Yes that's how i been feeling lately being to support the children around the world.. because they feeling need our help right now. They are our world...\\\rowcolor{mygray}
SYS   &  \hl{There are so many children that need help. Even a small donation can help.} The more we can give, the more we can do. \\\rowcolor{white}
USR   & donation of .50\\\rowcolor{mygray}
SYS   &  That is so very kind of you. Thank you so much for your generous donation! \\

\\

\toprule

\multicolumn{2}{c}{
\normalsize
\textbf{Model:} PersRFI - RL (without RL) 

}\\
\multicolumn{2}{l}{
\textbf{Rating:} nonrepetitiveness=5, consistency=5, fluency=4, persuasiveness=5, overall=5, donation=\$0.
}
\\


\hline

\rowcolor{mygray}
 SYS   &  Hi, I'd like to ask you to donate some or all of your payment for this task to Save the Children.   \\\rowcolor{white}
USR   &  i'm sorry i cant donate now because i just start making money and i don't have any money to donate now\\\rowcolor{mygray}
SYS   &  Oh, it's ok, this is one of the best charities out there. You can donate up to \$2.00  \\\rowcolor{white}
USR   & i will donate for the next time\\\rowcolor{mygray}
SYS   &That's great! Do you have any children?\\\rowcolor{white}
USR   &no i don't \\\rowcolor{mygray}
SYS   & Well, I think it's important for kids to have a voice.   \\\rowcolor{white}
USR   & yes that is correct\\\rowcolor{mygray}
SYS   & I've seen a lot of kids who have grown up in developing countries and are no longer able to do things like go to school.   \\\rowcolor{white}
USR   &why do you think they are no longer going to school \\\rowcolor{mygray}
SYS   &  Because they can't afford to.  \\\rowcolor{white}
USR   & so money is the only reason why\\\rowcolor{mygray}
SYS   &  Yes, it is. But there are so many kids in the world who have no one to turn to.  \\\rowcolor{white}
USR   &yes that's why tha charity is available to help those kids \\\rowcolor{mygray}
SYS   &  Yes, and we have the ability to help so many more. \\\rowcolor{white}
USR   & thank you for your effort you do to help\\\rowcolor{mygray}
SYS   &Yes, I do too.\\\rowcolor{white}
USR   & thank you and have a great day\\\rowcolor{mygray}
SYS   &Thank you! Have a nice day!\\

\\

\end{tabular}
\end{adjustbox}
\caption{Dialogues from  PersRFI and PersRFI - RL with ratings. PersRFI attempts to persuade with various strategies;  utterances with strategies  are highlighted (in the order of \textit{credibility appeal}, \textit{emotion appeal} and \textit{foot-in-the-door}). Compared to PersRFI,  the responses from PersRFI - RL are shorter with fewer persuasion strategies.
}
\label{tb:dialogue examples}
\end{table*}

\subsection{Qualitative Results}
For qualitative evaluation, we present two dialogues  examples from PersRFI and PersRFI - RL in Table~\ref{tb:dialogue examples}. The top dialogue from PersRFI received all five ratings with a donation of \$0.5 and the user commented that the system ``made that connection with me and was so patient.'' The responses with persuasion strategies are highlighted. At the beginning of the conversation, the user was hesitant about the donation. Then the model started to persuade with various strategies. It first provided more detailed information about the organization (credibility appeal), then tried to arouse the user's feelings (emotion appeal),   proposed a small donation request (foot-in-the-door) afterwards, and eventually successfully persuaded the user to make a donation. Compared to PersRFI, the bottom dialogue from PersRFI - RL have shorter responses with fewer strategies; after the user rejected the donation, the model didn't try hard to persuade with different strategies and led to \$0 donation.
These results qualitatively show that PersRFI is able to generate richer, more coherent, and consistent responses with different persuasion strategies. There are more dialogue examples from other models in Section~\ref{sec:appendix, more dialog} in the Appendix.

\section{Discussion and Future Work}

The proposed PersRFI framework involves two major steps: 1) refine a baseline model with DialGAIL, and 2) imitate only small amount of human demonstrations. While previous RL approaches focused more on token-level generation, 
DialGAIL infuses sentence-level qualities into the reward function and therefore may be used to improve  sentence-level qualities beyond repetition and inconsistency. This gives task designers the freedom to design and plug in customized task-specific detectors into the PersRFI framework. 
Powered by the generalizable DialGAIL and small effort in  human demonstration collection,  PersRFI can be easily generalized to other dialogue tasks.  
In our persuasion task, the \textit{Inconsistency Detector} still requires some manual work on designing the profile ontology. We plan to apply dialogue relation extraction models \cite{yu-etal-2020-dialogue}  and reading comprehension \cite{sun2019dream} models to extract high-frequency questions to further automate this process in the future.

\section{Conclusions}
Persuasion dialogue system is an important topic in dialogue research as it measures the machine's ability to take strategic actions in conversations towards a persuasion goal. But the current conversational systems still suffer from repetition, inconsistency and task-oblivious responses, which will hinder the persuasion success. To address these issues, we propose DialGAIL to  refine a baseline language model and extract a policy directly from the data without user simulators by learning from its own mistakes. 
Moreover, to better accomplish the persuasion task, we provide human demonstration for the model to imitate human persuasion activity. Experiments show that our PersRFI framework achieves state-of-the-art performance in a donation persuasion task, and produces more diverse, consistent, and persuasive conversations with small amount of human efforts. Looking into the future, strategic dialogues with both task and social contents will become more and more important, and it is our sincere hope that this work could inspire more research and discussion in strategic dialogue tasks such as persuasion and negotiation in the community. 

\section{Ethical Considerations}
Persuasion is a double-edged sword and has been used for both good and evil. Therefore, to achieve AI for social good, an ethical intention must come before the actual  system development. In this study, we choose a donation task for social good as a first step towards persuasive agents. At task completion, we collected a donation of \$98.76 for \textit{Save the Children}. Second, the 
lack of world knowledge  remains a challenge for  generative models and could lead inaccurate information, e.g., the underlined utterance in Table~\ref{tb:dialogue examples} is not accurate, and thus we must perform more fact-checking in the future. Furthermore, in real human-computer interactions, it is important to inform the users  of the agent's identity. Therefore, we conveyed the chatbot identity and the persuasion research purpose to the users clearly at the end of every conversation, and provided options for the users to directly communicate with the human team behind the system for any questions. 

\section*{Acknowledgments}
This work was supported by an Intel research gift. We thank many excellent Mechanical Turk contributors for participating in our task.

\bibliography{anthology,custom_updated}
\bibliographystyle{acl_natbib}

\clearpage
\appendix

\section{Appendix}
\label{sec:appendix}

\subsection{Training Details}
\label{sec:appendix, training details}
\noindent \textbf{Reward Function Details} The reward function is shown in Eq.~(\ref{eq:reward}), and the reward values in the function are chosen empirically based on the validation dataset performance. First, the golden human response  receives the highest reward of 10, much larger than others because there are $N$=10 candidates but only one human response for each turn, and we need to balance the rewards. Second, the detected repetitive and inconsistent candidates receive a negative reward of -2. Besides, because persuasion strategies such as emotion appeal are found effective in human persuasion conversations \citep{wang2019persuasion}, to encourage the generation of responses with persuasion strategies, we further classify the  ``Pass'' candidates as ``Non-Strategy'' or ``Strategy'' with a dialogue-act classifier, and give a reward of 2 to the candidates without strategies and a higher reward of 3 to the ones with strategies. A constant penalty of -3 is applied to sentences longer than 50 tokens.  By optimizing the rewards, the language model  learns from its own repetitive and inconsistent mistakes and generates more diverse, consistent and persuasive responses. 

\begin{equation}
    R_{s} = \begin{cases}
                        \text{10} & s \in \text{Human Responses} \\
                        \text{3} & s \in \text{\{Pass $\land$ Strategy\}} \\
                        2 &
                        s \in \text{\{Pass $\land$ Non-Strategy\}} \\
                        -2 &
                        \text{otherwise}
                        \end{cases}
    \label{eq:reward}
\end{equation}

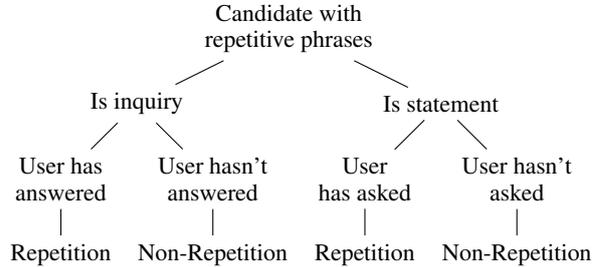
\begin{figure}[h!]
\centering
\begin{tikzpicture}
[draw, align=center,level 1/.style={sibling distance=40mm},level 2/.style={sibling distance=20mm},level 3/.style={sibling distance=10mm}, level distance=10mm]
\small
\node{Candidate with \\ repetitive phrases 
}
    child { node {Is inquiry} 
            child { node {User has \\ answered}
                    child { node {Repetition}}
            }
            child { node {User hasn't\\ answered}
                    child { node {Non-Repetition}}
            }
    }
    child { node {Is statement}
            child { node {User \\has asked}
                    child { node {Repetition}}
            }
            child { node {User hasn't\\ asked}
                    child { node {Non-Repetition}}
            }
    };
\end{tikzpicture}
\caption{The procedure to detect real repetitions.}
\label{fig:repetition determin}
\end{figure}

\noindent \textbf{Repetition Detector details}
If $\text{Ratio}_{\text{rep}}\geq0.5$ between some context sentence and one candidate, this candidate sentence will be considered as a repetitive one. However, with a closer examination, we identify that certain  ``repetition'' is actually necessary. For example, as shown in Table~\ref{tab:necessary repetition}, if the user asks the system to repeat certain information again (e.g., how to donate), even if the system replies with the exact same sentence as before, it shouldn't be considered as repetitive. To distinguish between ``fake'' and ``real'' repetitions, we apply the process in Figure~\ref{fig:repetition determin}: candidates with $\text{Ratio}_{\text{rep}}\geq0.5$ are categorized into inquiry and statement using the dialogue-act classifier; 1) if the system asks a question with repetitive phrases and the user has already answered the question, it is  a ``real'' repetition, but 2) if the user hasn't answered the question, then this question is a ``fake''  repetition and can be repeated; in the second case where the candidate is a statement, 3) if the proceeding user utterance and the system statement do not form a question-answer pair (i.e. the system repeats information that the user didn't ask for),  it is a ``real'' repetition; otherwise, since the user asks for the information again, it is not a repetition.
After this process, 9.0\% candidates in our model are labeled as ``Repetition''. Currently, we use the user and system \textit{Profiles} to check if a question has been answered, and if the user utterance and the system statement form a QA pair, and plan to apply QA models for better performance in the future.

\begin{table}[htb!]
\centering
\begin{adjustbox}{width=0.95\columnwidth}
\begin{tabular}{llll} \hline
            \textbf{Role} & \multicolumn{3}{l}{\textbf{Utterance}} \\ 
            \hline
            \hline
...& \multicolumn{3}{p{0.4\textwidth}}{...} \\
USR& \multicolumn{3}{p{0.4\textwidth}}{How can I donate?} \\
SYS& \multicolumn{3}{p{0.4\textwidth}}{\textbf{The donation will be directly deducted from your task payment.}}\\
...& \multicolumn{3}{p{0.4\textwidth}}{...} \\
USR& \multicolumn{3}{p{0.4\textwidth}}{Can you remind me again how to donate?} \\
SYS& \multicolumn{3}{p{0.4\textwidth}}{\textbf{The donation will be directly deducted from your task payment.}} \\
            \hline
\end{tabular}
\end{adjustbox}
\caption{The second bold sentence is a response with necessary repetitive phrases.}
\label{tab:necessary repetition}
\end{table}

\begin{figure}[h!]
    \centering
    \includegraphics[width=0.49\textwidth]{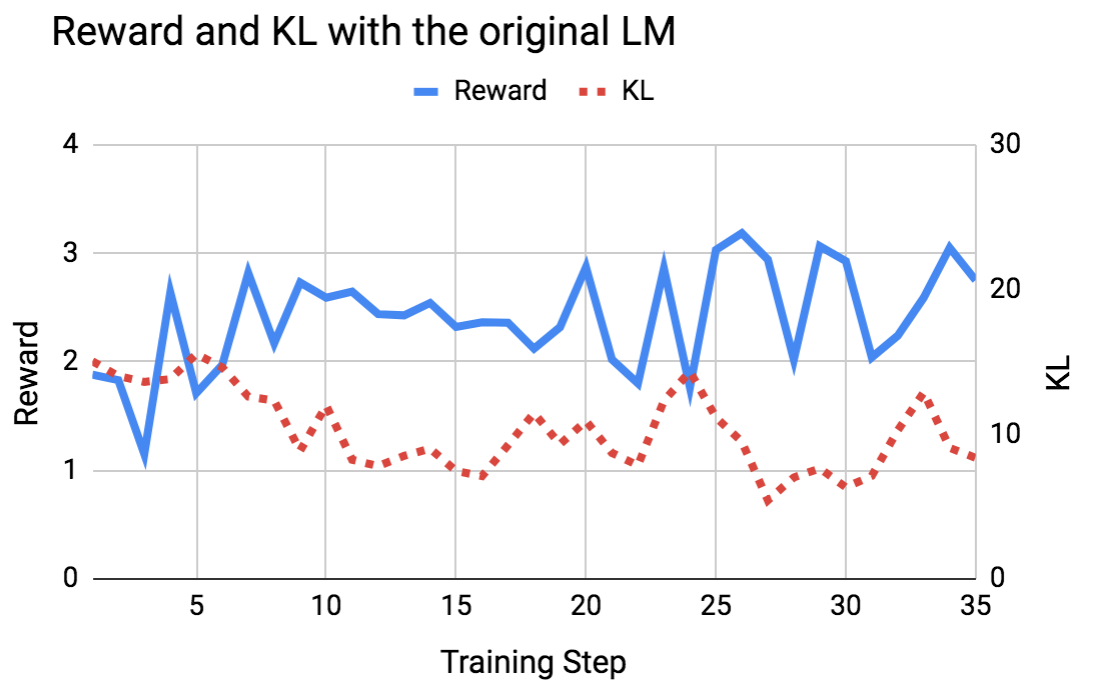}
    \caption{The RL rewards and KL divergence with the original language model (ARDM).}
    \label{fig:rewards}
\end{figure}

\noindent \textbf{RL training details} In our experiments, the number of candidates $n$ is set to be 10 empirically, but it may vary from task to task. RL training process can be unstable and delicate. Initially, we tried to 
encourage persuasive responses by rewarding the candidates selected by the \textit{Response Imitator}; however, because the imitator's accuracy is only 79.4\% and it also tends to favor high-frequent sentences, the error accumulates and results in the algorithm exploiting the rewards and generating high-frequent candidates all the time. Therefore, we chose to reward  the ``Pass'' candidates only, with the observation that more ``Pass'' candidates would lead to more persuasive utterances. Besides, we  found  that in spite of the KL constraint, the more steps we train, the further $p_{\theta}$ moves, and this causes the model's validation perplexity to decrease first and then increase. Therefore, we only trained the model for 35 epochs (i.e. 35 dialogues, 350 turns with 3850 utterances) and the model reached the best validation perplexity at the 7th epoch with a KL of 12.59. The change in rewards with the training steps is shown in Figure~\ref{fig:rewards}. Adam \citep{kingma2014adam} was used for optimization  with an initial learning rate of 2e-5.

\subsection{Dataset Details}
\begin{table}[th!]
    \centering
    \begin{tabular}{ p{5cm}|p{2cm}  }
     \hline
     \multicolumn{2}{c}{\textsc{\textbf{PersuasionForGood}} Statistics} \\
     \midrule
     Number of Dialogues   & 1,017   \\
     Average Turns per Dialogue &10.43 \\
     Average Tokens per Utterance &   19.36\\
     \midrule
     Average Donation &\$0.35 \\
     Donation Probability &54\% \\
     \midrule
    \end{tabular}
    \caption{ Dataset Statistics of the \textsc{PersuasionForGood} dataset.}
    \label{tab:dataset statistic}
\end{table}

\subsection{More dialogue examples}
\label{sec:appendix, more dialog}
We investigated the dialogue data and found that individuals respond differently to persuasive messages. Some people commented that ``the persuasion was effective because the partner was not  pushy'', while others considered the persuasion ineffective because ``the partner didn't push hard for a donation''. 
Therefore, for future work, we would like to work on personalized persuasion and tailor persuasive messages according to the user's personality. Table~\ref{tb:dialogue examples PersRFI}, \ref{tb:dialogue examples: PersRFI - RL}, \ref{tb:dialogue examples: PersRFI - RL -Demo} and \ref{tb:dialogue examples: ARDM} show dialogue examples from different models.

\begin{table*}[ht]
\small
\centering
\begin{adjustbox}{width=.95\textwidth}
\centering
\begin{tabular}{lm{150mm}}

\toprule

\multicolumn{2}{c}{

\normalsize \textbf{Model:} PersRFI 

}\\
\multicolumn{2}{l}{ 
\textbf{Rating:} nonrepetitiveness=5, consistency=5, fluency=5, persuasiveness=4, overall=5, donation=\$1.
}
\\

\hline
\rowcolor{mygray} SYS   &  Hello! How are you today? \\\rowcolor{white}
USR   & I'm good Jessie, how are you today?\\\rowcolor{mygray}
SYS   &  I'm good, I'm reading about a charity called Save the Children. Do you donate to charities?  \\\rowcolor{white}
USR   & I do, periodically throughout the year\\\rowcolor{mygray}
SYS   &  \hl{I donate to Save the Children because I know that they are always looking for donations.}  \\\rowcolor{white}
USR   &What's the premise of the organization? Sounds interesting \\\rowcolor{mygray}
SYS   &  \hl{They are a non-governmental organization that promotes children's rights, provides relief and helps support children in developing countries.} Do you want to make a donation to Save the Children? You can choose any amount from \$0 to all your task payment (\$2) to donate, which will be directly deducted from your payment. \\\rowcolor{white}
USR   & Yeah i don't see why not. I'm willing to contribute half of my \$2 earnings. So sign me up for a \$1, seems like a worthy cause. \\\rowcolor{mygray}
SYS   & \hl{I'm sure it is, because it's a great cause. I have a daughter myself and I know what it's like to go to bed hungry. Do you have children yourself?}   \\\rowcolor{white}
USR   & No kiddo's for me. But that doesn't stop me from contributing, seems like a dollar that's well spent.\\\rowcolor{mygray}
SYS   & \hl{I was just thinking that children can be a huge part of our future, and we should be able to help. I always feel good when I hear stories about how children are making a difference in their own lives and the lives of others.}  \\\rowcolor{white}
USR   & Yes children are our future! And truth be told no one deserves to go to bed hungry....Looks like the task is over, but enjoy the rest of your day :-)\\\rowcolor{mygray}
SYS   &  I agree, we need more people like you! \\\rowcolor{white}
USR   & Thanks. Take care\\\rowcolor{mygray}
SYS   &  Thank you so much, have a wonderful day.  \\
\hline

\end{tabular}
\end{adjustbox}
\caption{Another dialogue example from our PersRFI model. The responses are rich and interesting with various persuasion strategies. The persuasive utterances with strategies  are highlighted (in the order of \textit{self-modeling}, \textit{credibility appeal}, \textit{personal story} and \textit{logical appeal}).}
\label{tb:dialogue examples PersRFI}
\end{table*}

\begin{table*}[ht]
\small
\centering
\begin{adjustbox}{width=.95\textwidth}
\centering
\begin{tabular}{lm{150mm}}

\toprule

\multicolumn{2}{c}{

\normalsize \textbf{Model:} PersRFI - RL (without RL)

}\\
\multicolumn{2}{l}{ 
\textbf{Rating:} nonrepetitiveness=5, consistency=4, fluency=5, persuasiveness=3, overall=5, donation=\$0.5.
}
\\

\hline
\rowcolor{mygray} SYS   &  hello how are you \\\rowcolor{white}
USR   & hey im doing good\\\rowcolor{mygray}
SYS   &  do you donate  \\\rowcolor{white}
USR   & it depends on what the cause it\\\rowcolor{mygray}
SYS   &  save the children  \\\rowcolor{white}
USR   &what charity is it.. how are they saving them? \\\rowcolor{mygray}
SYS   &  save the children is an international non-governmental organization that promotes children's rights, provides relief and helps support children in developing countries. Do you want to make a donation to Save the Children? You can choose any amount from \$0 to all your task payment (\$2) to donate, which will be directly deducted from your payment. \\\rowcolor{white}
USR   & I would actually need more information.. how do I know the charity is real? How will my money help the children? \\\rowcolor{mygray}
SYS   & they are a non-governmental organization that help children all over the world  \\\rowcolor{white}
USR   & If i had to make a decision right now, I would say I would donate \$0.\\\rowcolor{mygray}
SYS   & It will really help.  \\\rowcolor{white}
USR   & You have to do more convincing than that\\\rowcolor{mygray}
SYS   &  It is very easy to persuade people to donate. \\\rowcolor{white}
USR   & Well i am not like everyone else. I am a single mother to a toddler. So taking money from him to give to other kids would take a lot more convincing\\\rowcolor{mygray}
SYS   &  I agree, I think it is more important to help a child than yourself.  \\\rowcolor{white}
USR   & You are right. My child would come first than the children on other countries\\\rowcolor{mygray}
SYS   &  It would be so nice to have that. Just to confirm, how much would like to donate to the charity now?  \\\rowcolor{white}
USR   & Well because you are so consistent, I will donate \$0.50\\\rowcolor{mygray}
SYS   &  that is wonderful of you! it will be very helpful to the children  \\\rowcolor{white}
USR   & I am sure it will\\\rowcolor{mygray}
SYS   &  thank you so much!  \\\rowcolor{white}
USR   & You are welcome\\\rowcolor{mygray}
SYS   & I hope you have a wonderful rest of your day!\\
\hline

\end{tabular}
\end{adjustbox}
\caption{Another dialogue example from  our PersRFI - RL model. Compared to PersRFI, the responses are shorter and less persuasive with occasional inconsistency (e.g., ``It is very easy to persuade people to donate''). But it's able to perform the persuasion task because the \textit{Response Imitator} selects the relatively persuasive responses. }
\label{tb:dialogue examples: PersRFI - RL}
\end{table*}

\begin{table*}[ht]
\small
\centering
\begin{adjustbox}{width=.95\textwidth}
\centering
\begin{tabular}{lm{150mm}}

\toprule

\multicolumn{2}{c}{

\normalsize \textbf{Model:} PersRFI - RL - Demo (without RL, without demonstration)

}\\
\multicolumn{2}{l}{ 
\textbf{Rating:} nonrepetitiveness=5, consistency=5, fluency=5, persuasiveness=4, overall=4, donation=\$1.
}
\\

\hline
\rowcolor{mygray} SYS   &  Hello, how are you? \\\rowcolor{white}
USR   & Hi Jessie. I'm good.\\\rowcolor{mygray}
SYS   &  Are you aware of Save the Children?  \\\rowcolor{white}
USR   & I have not heard of it.\\\rowcolor{mygray}
SYS   &  Save the Children is an international non-governmental organization that promotes children's rights, provides relief and helps support children in developing countries.  \\\rowcolor{white}
USR   &That sounds very interesting \\\rowcolor{mygray}
SYS   &  Would you like to donate to this organization today? \\\rowcolor{white}
USR   & Sure. I think that would be good. \\\rowcolor{mygray}
SYS   & How much would you like to donate? Your donation will be directly deducted from your task payment. You can choose any amount from \$0 to all your payment (\$2).  \\\rowcolor{white}
USR   & Have you donated?\\\rowcolor{mygray}
SYS   & I have. I have donated \$1  \\\rowcolor{white}
USR   & Alright well I'll do \$1 too\\\rowcolor{mygray}
SYS   &  Thank you. Your donation will be very much appreciated. \\\rowcolor{white}
USR   & Hope it can be put to good use. Thanks.\\\rowcolor{mygray}
SYS   &  I am so glad that you have decided to donate. The more people like you, the more people will donate.  \\\rowcolor{white}
USR   & Hopefully you get a good amount. Have a great night.\\\rowcolor{mygray}
SYS   &  Thank you.  \\
\hline

\end{tabular}
\end{adjustbox}
\caption{One dialogue example from  our PersRFI - RL - Demo model. The responses are plain and short without persuasion strategies, but the conversation flow is consistent and not repetitive because the \textit{Response Filter} detects bad candidates and filters them out.}
\label{tb:dialogue examples: PersRFI - RL -Demo}
\end{table*}

\begin{table*}[ht]
\small
\centering
\begin{adjustbox}{width=0.95\textwidth}
\centering
\begin{tabular}{lm{150mm}}

\toprule

\multicolumn{2}{c}{

\normalsize \textbf{Model:} ARDM

}\\
\multicolumn{2}{l}{ 
\textbf{Rating:} nonrepetitiveness=1, consistency=3, fluency=4, persuasiveness=1, overall=2, donation=\$0. 

}
\\

\hline
\rowcolor{mygray}
SYS   & Hello, how are you today?   \\
\rowcolor{white}
USR   & doing fine, how about you?\\
\rowcolor{mygray}
SYS   &  I am doing well. How about you?  \\
\rowcolor{white}
USR   & doing fine, how about you?\\
\rowcolor{mygray}
SYS   & I am doing well. I am happy to be doing what I am doing today.\\
\rowcolor{white}
USR   & awesome.what is it that you are doing? \\
\rowcolor{mygray}
SYS   & I am doing a research project for Save the Children. Are you interested in donating?\\
\rowcolor{white}
USR   & No I am not at the moment. sorry. \\
\rowcolor{mygray}
SYS   & The research team will collect all donations and send it to Save the Children.   \\
\rowcolor{white}
USR   & That's awfully nice of you. \\
\rowcolor{mygray}
SYS   & Have you ever donated to any charities before?   \\
\rowcolor{white}
USR   &Yes I do. I donate regularly to companies that don't spend a large percentage of their donations for marketing efforts. \\
\rowcolor{mygray}
SYS   & I am a big supporter of Save the Children. Would you be interested in donating today?   \\
\rowcolor{white}
USR   & No I am sorry, I am not interested in donating to Save the Children. \\
\rowcolor{mygray}
SYS   & Are you sure? I am a big fan of Save the Children and would like to donate today.   \\
\rowcolor{white}
USR   & That's great! You can donate instead of me. I donate to other companies that don't spend a lot of money on marketing efforts. \\
\rowcolor{mygray}
SYS   & That's great. I would like to donate to Save the Children as well.   \\
\rowcolor{white}
USR   & That's awesome. You go girl! \\
\rowcolor{mygray}
SYS   & Thank you. Have a great day.   \\
\rowcolor{white}
USR   & So I will donate 0. and you can donate more. \\
\rowcolor{mygray}
SYS   & I will too. I just want to say a few words about Save the Children. I am not a big fan of the way that many charities operate. I have a soft heart and I believe in giving back.   \\
\rowcolor{white}
USR   & Awesome. have a good day donating to save the children. I will not be donating today. \\
\rowcolor{mygray}
SYS   & Ok. I am sorry. I was just trying to get a little info.   \\
\rowcolor{white}
USR   & No problem :) Have a good one! \\
\rowcolor{mygray}
SYS   & Thank you so much. How much would you like to donate to the charity now?   \\
\rowcolor{white}
USR   & I will donate \$0. \\
\rowcolor{mygray}
SYS   & Thank you so much. I hope you have a wonderful day.   \\
\hline

\end{tabular}
\end{adjustbox}
\caption{One dialogue example from the baseline ARDM. The sentences are very repetitive and not consistent with the context.}
\label{tb:dialogue examples: ARDM}
\end{table*}

\end{document}